\documentclass{article}
\usepackage{amsmath,graphicx}
\usepackage[preprint]{spconf}
\usepackage{textcomp,float}
\usepackage[numbers,sort&compress]{natbib}
\usepackage{tablefootnote}
\usepackage{booktabs}
\usepackage{multirow}
\usepackage{stfloats}

\title{Climate and Weather:
Inspecting Depression Detection \\ via Emotion Recognition}

\name{Wen Wu$^1$, Mengyue Wu$^2$$^{\dag}$, Kai Yu$^2$$^{\dag}$\thanks{$\dag$ Mengyue Wu and Kai Yu are the corresponding authors.}}
\address{
$^1$Cambridge University Engineering Department, Cambridge, UK\\
$^2$MoE Key Lab of Artificial Intelligence\\
X-LANCE Lab, Dept. of Computer Science and Engineering\\
AI Institute, Shanghai Jiao Tong University, Shanghai, China\\
\small{\texttt{ww368@cam.ac.uk; \{mengyuewu,kai.yu\}@sjtu.edu.cn}}}

\begin{document}
\copyrightnotice{\copyright\ IEEE 2022}

\maketitle
\begin{abstract}
Automatic depression detection has attracted increasing amount of attention but remains a challenging task. Psychological research suggests that depressive mood is closely related with emotion expression and perception, which motivates the investigation of whether knowledge of emotion recognition can be transferred for depression detection. This paper uses pretrained features extracted from the emotion recognition model for depression detection, further fuses emotion modality with audio and text to form multimodal depression detection. 
The proposed emotion transfer improves depression detection performance on DAIC-WOZ as well as increases the training stability. The analysis of how the emotion expressed by depressed individuals is further perceived provides clues for further understanding of the relationship between depression and emotion.

\end{abstract}
\section{Introduction}
\label{sec:intro}
Major Depressive Disorder (MDD) is a disease of great concern that affects more than 264 million people worldwide~\cite{GBD2017}. 
Psychology studies suggest that depressed mood can directly influence individual's emotion expression and perception~\cite{Cognitive}. Cognitive biases and deficits caused by MDD affect emotion regulation ability such as habitually increasing the use of emotion regulation strategies that serve to down-regulate positive emotion and reducing the use of strategies that serve to up-regulate positive emotion~\cite{VANDERLIND2020101826}.
Such a relation can be analogized as climate and weather: depression is a long-term mood that affects transient emotional expressions. 

Inspired by the relation between emotion and depression, two related machine learning tasks can be transferred -- knowledge from emotion recognition may be lent for depression detection.  
Depression detection aims at predicting one's mental state (depressed or healthy) from audio or text while emotion recognition outputs discrete emotional classes (e.g., happy, angry, frustrated) or continuous attribute scores (e.g., valence, activation, dominance).
Similarities lie between these two tasks such as the audio and text are the main modalities used to predict one's emotion category or depressive state. 

Though datasets on emotion and depression are generally small in size due to difficulties associated with data acquisition and costs of annotations, data sparsity is more servere for depression dataset since clinical interview is a sparse scenario. Compared with emotions, short-termed and transitional, depression is a more longitudinal mental state. 
Emotions are usually annotated at segment-level (i.e., one label per sentence) while the depression diagnosis decisions are usually made at session-level (i.e., one label per interview). Given the same size of data, the number of effective samples for depression detection task is smaller.

Prior work~\cite{Stasak_2016} investigated emotional speech in depression detection but relied on a rather small dataset with continuous affect ratings and depression scores. This paper uses larger emotion datasets for better generalization capability and presents robust depression detection results by using emotion features extracted from a well-pretrained emotion model. Furthermore, this paper aims at investigating the interaction between emotion and depression, looking at how transient emotion expression (weather) reflect longtidutinal depression (climate) and how longtidutinal depression (climate) can influence transient emotion expression (weather). By analyzing the emotional content of the depression data, this paper provides clues for understanding the relationship between depression and emotion.

\section{Transfer Emotion to Depression}
\label{sec: methods}
\begin{figure}[]
  \centering
  \includegraphics[width=0.9\linewidth]{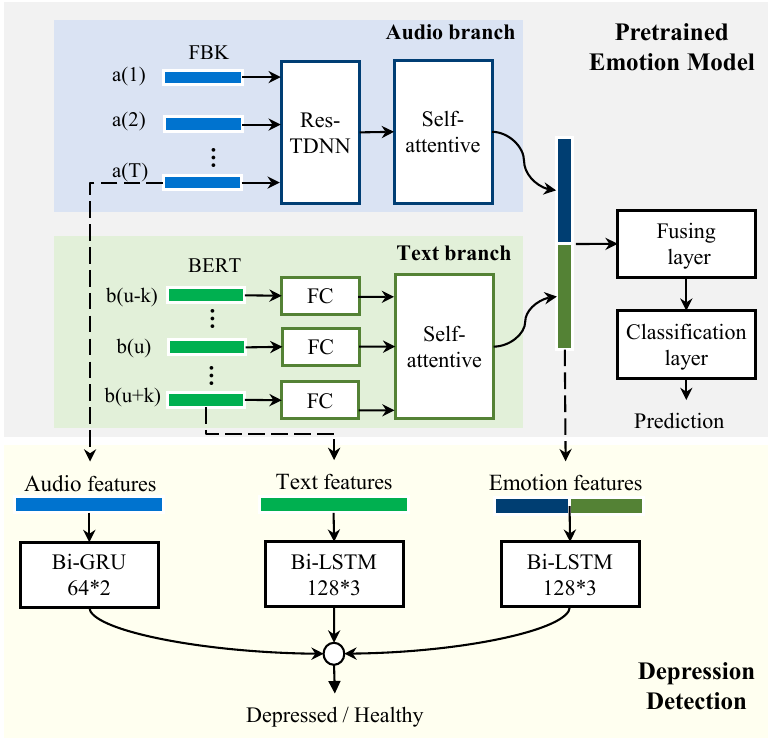}
    \caption{Process of transferring knowledge from emotion recognition to depression detection. An emotion recognition model is first trained. The outputs of its hidden layer are extracted as emotion features and used as the input to a depression detection system. Emotion modality is then fused with audio and text modalities to form multimodal depression detection.}
    \label{fig: process}
\end{figure}

The overall process of transferring knowledge from emotion recognition to depression detection is illustrated in Fig.~\ref{fig: process}. We first introduce the features used in the experiments, followed by our pretrained emotion model and the knowledge transfer approach in detail. 

\subsection{Feature representation}
Spectral based features have consistently been observed to change with a speaker’s mental state~\cite{audio_review}. Log Mel filterbank features (FBKs) have been commonly used as audio representation for speech-based emotion recognition \cite{Metallinou_2010, Kim_2013, Wu_2021} and depression detection \cite{Ma_2016,Rejaibi2019ClinicalDA}. In this paper, 40-dimensional (-d) FBKs with a 10 ms frame 
shift and 25 ms frame length along with first derivative coefficients are used as audio features.

The use of large unsupervised pretrained language models have achieved excellent performance on a variety of language tasks.
Pretrained sentence-level embeddings derived from a large pretrained language model, such as BERT \cite{devlin-etal-2019-bert}, have drawn much attention. In this paper, we use the pretrained BERT-base model to encode the transcription of each single utterance into a 768-d vector.

\subsection{Pretraining: emotion recognition}
\label{sec: model structure}
The emotion model proposed in \cite{Wu_2021} is adapted for emotion recognition pretraining, which consists of an audio branch and a text branch.

In the audio branch, a time delay neural network with residual connections (Res-TDNN) \cite{Kreyssig2018ImprovedTU} is used as the encoder to derive the frame-level vectors, which consists of four Res-TDNN layers with layer width 512 and context width [2,+2], \{-1,+2\},\{-3,+3\},\{-7,+2\}. We then utilize a five-head self-attentive layer \cite{Lin2017ASS} to pool the frame-level vectors across time in the input window. A modified penalty term is used for the five-head self-attentive layer, with three heads set to obtain more ``spiky'' attention weight distributions and the other two set to obtain  ``smoother'' distributions \cite{Sun_2019}. 

The text branch focuses on capturing meaning from the speech transcriptions. The embeddings for consecutive utterances of context [-3,+3] (three in the past, current, and three in the future) are used as the input since the emotion of each utterance is often strongly related to its context in a spoken dialogue~\cite{Mark_2012}.
In the text branch, a shared fully-connected (FC) layer is used to reduce the dimension of each input sentence embedding, and the resulting vectors are then integrated by another five-head self-attentive layer, whose attention-weight distribution reflects the extent to which sentences in the context affect the current emotion. 

During pretraining, the output of the audio branch (128-d) and the text branch (320-d) are then concatenated and fed into an FC layer for fusion, followed by the classification layer.

\subsection{Depression detection with emotion transfer}

The parameters of the pretrained emotion model are then fixed and the pretrained model is treated as an emotion feature extractor. The emotion features are obtained at the input to the fusing layer as shown in Fig.~\ref{fig: process}. As mentioned in Section~\ref{sec:intro}, emotion recognition is usually conducted at segment-level and the emotion features are thus extracted at segment-level. An additional bi-directional Long Short-Term Memory (bi-LSTM) model is then trained to pool the segment-level features and produce session-level decision of depression diagnosis.

Then, emotion is seen as a separate modality besides audio and text. The system above serves as a unimodal system for emotion modality.

Audio and text modality are trained using the same features as in pretraining the emotion model (FBKs and BERTs), with audio trained through a bi-directional Gate Recurrent Unit (bi-GRU) and text through a bi-LSTM model.
As shown in Fig.~\ref{fig: process}, three modalities are then fused for multimodal depression detection.

\section{Experiments and Results}
\label{sec: experiment}
We use separate datasets for experiments on emotion recognition and depression detection, described in Section~\ref{sec: dataset}. Section~\ref{sec: emo exp} demonstrates the pretrained emotion recognition models with different configurations. Emotion features extracted from different pretrained models are then used for depression detection, results compared in Section~\ref{sec: depression detection}. The best performing emotion features then represent the emotion modality to fuse with audio and text modality in Section~\ref{sec:fusion}.

\subsection{Datasets}
\label{sec: dataset}
The emotion datsets involved are IEMOCAP~\cite{Busso2008IEMOCAPIE} and CMU-MOSEI~\cite{mosei}. We experiment on these two datasets for a comprehensive interaction with depression, as they represent different emotion aspects, i.e. categorical label and sentiment analysis.

Emotion features extracted from the emotion recognition model pretrained on these two datasets are compared and the best performing emotion features are selected as the emotion modality for fusion on depression detection.

\textbf{IEMOCAP}: The IEMOCAP corpus is a multimodal dyadic conversational dataset. It  contains a total of 5 sessions and 10 different speakers, with a session being a conversation of two exclusive speakers. IEMOCAP provides both discrete categorical based annotations, and continuous attribute based annotations. To be consistent with previous studies, only utterances with ground truth labels belonging to ``angry", ``happy", ``excited", ``sad", and ``neutral" were used. The ``excited'' class was merged with ``happy'' to better balance the size of each emotion class.

\textbf{CMU-MOSEI}: CMU-MOSEI is a multimodal sentiment and emotion analysis dataset made up of 23,454 movie review video clips taken from YouTube. Each sample is labeled by human annotators with a sentiment score from -3 (strongly negative) to 3 (strongly positive). 

~\\
The downstream depression detection experiment is conducted on a benchmark dataset DAIC-WOZ~\cite{DAIC-WOZ}.

\textbf{DAIC-WOZ}: DAIC-WOZ dataset is a commonly used dataset within depression detection which encompasses 50 hours of data collected from  189 clinical interviews from a total of 142 patients. This corpus is created from semi-structured clinical interviews. 
Thirty speakers within the training (28 \%) and 12 within the development (34 \%) set are classified to have depression. Following prior works, results in our paper are reported on the development subset.

\subsection{Pretrained emotion recognition}
\label{sec: emo exp}
IEMOCAP-based model was trained using multitask learning. A combination of cross entropy and Huber loss was used for classification of categorical labels and regression of continuous labels respectively. Both the weighted accuracy (WA) and unweighted accuracy (UA) are reported for classification. Mean absolute error (MAE) is reported for regression. MOSEI-based model was trained by binary sentiment classification. The model performance was evaluated using binary accuracy (Acc2) and F1 score. 

The use of audio features, text features and their combination were investigated for both datasets and the results are shown in Table~\ref{tab: pretrained model}. The systems were well-trained compared with the SOTA results\footnote{SOTA classification accuracy on IEMOCAP is around 77\% for fused system \cite{Poria_2018,Yoon_2019,ijcai2019-703}. The results cannot be compared directly due to inconsistent selection of test set. SOTA results on MOSEI vary from 80 to 84\% \cite{tsai_2019-multimodal,Sun_2020}.}.
 \begin{table}[htbp!]
\centering

\begin{minipage}[b]{1.0\linewidth}
  \centering
  \scalebox{0.9}{
  \begin{tabular}{c|cccccc}
\toprule
&Mod. & WA     & UA     & MAE-v  & MAE-a  & MAE-d  \\
\midrule
\multirow{3}{*}{IEMOCAP} & A    & 0.6488 & 0.6656 & 0.6182 & 0.4729 & 0.6225 \\
& T     & 0.7326 & 0.7350  & 0.4848 & 0.5113 & 0.6161 \\
& A+T     & 0.7718 & 0.7787 & 0.4501 & 0.4404 & 0.559 \\
\bottomrule
\end{tabular}
    }
  \centerline{(a) Emotion recognition results on IEMOCAP.}
 \vspace{1mm}
\end{minipage}

\begin{minipage}[b]{1.0\linewidth}
  \centering
  \scalebox{0.9}{
  \begin{tabular}{c|ccc}
\toprule
& Mod. & Acc2    & F1     \\
\midrule
\multirow{3}{*}{MOSEI} & A    & 0.7239 & 0.8060 \\
& T     & 0.7909 & 0.8314 \\
& A+T     & 0.7953 & 0.8361 \\
\bottomrule
\end{tabular}
}
  \centerline{(b) Emotion recognition results on CMU-MOSEI.}
\end{minipage}

 \vspace{-2mm}
\caption{Emotion recognition results of pretrained models. Mod. denotes modality present from \{(A)udio, (T)ext\}.}
\label{tab: pretrained model}
\end{table}
 \vspace{-2mm}

\subsection{Depression detection using extracted emotion features}
\label{sec: depression detection}
Emotion features were respectively extracted from the pretrained models in Table~\ref{tab: pretrained model} and fed as the input for depression detection on DAIC-WOZ, corresponding results compared in Table~\ref{tab: emo-dep}. 
To reduce the impact of initialization on sparse datasets, experiments were run for 20 different seeds and both the average (AVG) and the best (MAX) are reported along with the standard deviation (STD) across seeds. 
Overall, features extracted from MOSEI-based models produce better results than those from IEMOCAP-based models.
We speculate this relates with the fact that CMU-MOSEI is a larger dataset than IEMOCAP (66 hours vs 12 hours) and contains much more speakers (1000 vs 10), resulting in stronger generalization potential.
Further, IEMOCAP is an acted dataset extravagantly conveying emotions while CMU-MOSEI includes excerpts with emotions naturally expressed, exhibiting a similar format to DAIC-WOZ.

Merging audio and text modalities improves the emotion classification results by $\sim$ 4\% for IEMOCAP-based model (see Table~\ref{tab: pretrained model}(a)), but leads to a decrease in depression detection performance. 
The discrepancies between Table~\ref{tab: pretrained model} and Table~\ref{tab: emo-dep} indicates that although emotion information is useful in depression detection, higher emotion classification results do not necessarily correspond to better depression detection performance. Besides, for both IEMOCAP and MOSEI, features extracted from emotion model pretrained using only audio information (abbr. audio-based emotion features) yields poor performance, which will be further discussed in Section~\ref{sec: analysis}.

\begin{table}[htbp!]
\centering
\scalebox{0.9}{
\begin{tabular}{c|cccc}
\toprule
 & Mod. & F1(MAX) & F1(AVG) & F1(STD) \\
\midrule
\multirow{3}{*}{IEMOCAP} & A                     & 0.558   & 0.529   & 0.018  \\
&   T                      & 0.621   & 0.562   & 0.022  \\
&   A+T                      & 0.600     & 0.554   & 0.025 \\
\midrule
\multirow{3}{*}{MOSEI} & A                     & 0.558   & 0.534  & 0.014  \\
&   T                      & 0.750    & 0.718  & 0.018 \\
&   A+T                      & 0.828   & 0.771   & 0.027\\
\bottomrule
\end{tabular}
}
\caption{Depression detection results using emotion features extracted from pretrained emotion models in Table~\ref{tab: pretrained model}. Here ``A" and ``T" denotes the modality used in pretraining the emotion recognition model.}
\label{tab: emo-dep}
\end{table}

\subsection{Fusion}
\label{sec:fusion}
We then treat emotion as an individual modality. 
The best performing emotion features were selected as the input to the emotion modality, which is the last row in Table~\ref{tab: emo-dep} extracted from MOSEI-based model.
Emotion modality is then fused with audio and text modalities, shown in Table \ref{tab: dep-fusion}.
Again, the experiments were run for 20 seeds.
Comparing Table~\ref{tab: emo-dep} and Table~\ref{tab: dep-fusion}, emotion features of various configuration all produce much smaller standard deviation than directly using audio or text features, suggesting improved stability and robustness of the system against data sparsity. 
An interesting observation stems from the results that direct audio features outperform audio-based emotion features. Some useful information for depression detection may have been discarded when extracting emotion features from the audio data.
Late fusion based on voting mechanism produces an F1 of 0.869, which is in line with SOTA depression detection results on DAIC-WOZ.
A pattern can be observed that when facing conflicted votes, accuracy is guaranteed to follow the emotion modality. 

 \vspace{-2mm}
\begin{table}[htbp!]
\centering
\scalebox{0.9}{
\begin{tabular}{c|ccc}
\toprule
Modality & F1(MAX) & F1(AVG) & F1(STD) \\
\midrule
Direct Audio           &          0.696   & 0.608  & 0.033  \\
Direct Text                      & 0.727    & 0.600  & 0.070 \\
 Emotion                      & 0.828   & 0.771   & 0.027\\
\midrule
Fused  &   \multicolumn{3}{c}{0.869}\\
\bottomrule
\end{tabular}
}
\caption{Depression detection results of three modalities.}
\label{tab: dep-fusion}
\end{table}
 \vspace{-2mm}

\section{Emotion analysis of depression data}
\label{sec: analysis}
\begin{figure*}[ht]

\begin{minipage}[b]{1.0\linewidth}
  \centering
  \includegraphics[width=0.9\columnwidth]{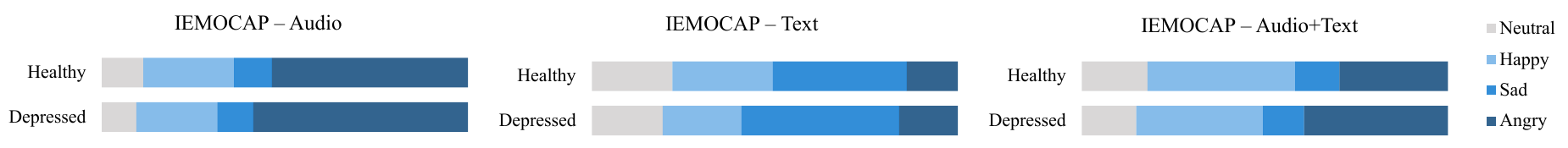}
  \centerline{(a) Categorical emotion distribution of depressed data.}\medskip
\end{minipage}
\begin{minipage}[b]{1.0\linewidth}
  \centering
  \includegraphics[width=0.9\columnwidth]{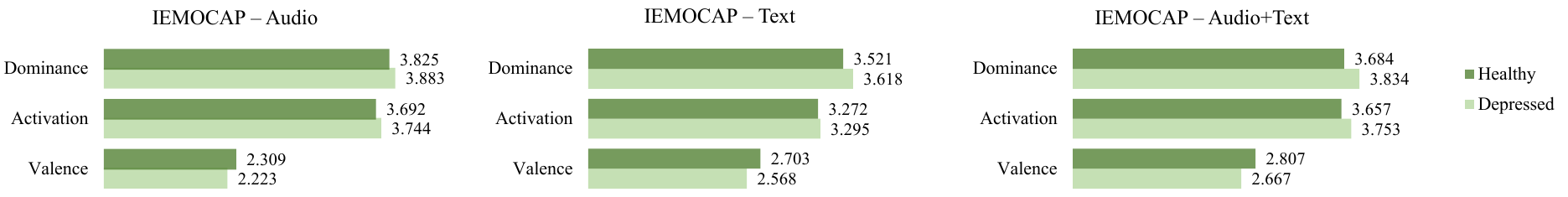}
  \centerline{(b) Attribute predictions of depressed data.}
 \vspace{-6mm}
\end{minipage}
    \caption{Emotion distribution of depressed data. Model trained on IEMOCAP.}
     \vspace{-2mm}
    \label{fig: ieomcap-cat+reg}
\end{figure*}

\begin{figure}[htbp!]
    \centering
    \includegraphics[width=\columnwidth]{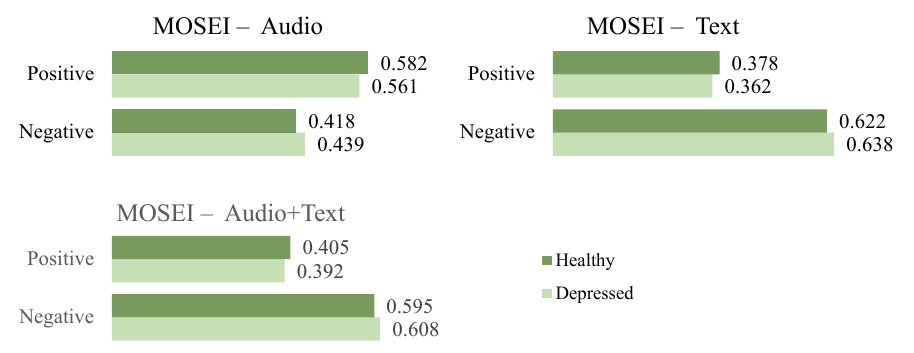}
    \caption{Binary sentiment prediction of depressed data. Model trained on CMU-MOSEI. }
     \vspace{-4mm}
    \label{fig: mosei-sent}
\end{figure}

This section analyzes the emotional content of depression data through different modalities. DAIC training set was fed into the trained emotion model in Section~\ref{sec: emo exp}. Outputs were averaged across all input samples. Results predicted by IEMOCAP- and MOSEI-based models are respectively shown in Fig.~\ref{fig: ieomcap-cat+reg} and Fig.~\ref{fig: mosei-sent}.

Fig.~\ref{fig: ieomcap-cat+reg}(a) illustrates the average classification probability score over four discrete emotion categories, where the use of different modalities yields inconsistent emotional distributions. The dominant emotion is \textit{angry} for the audio-alone system and \textit{sad} for the text-alone system. \textit{Happy} has the second highest score in both models and dominates the fused model. For all modalities, depressed samples have higher score in \textit{angry} and \textit{sad} while healthy samples have higher score in \textit{neutral} and \textit{happy}. 

Fig.~\ref{fig: ieomcap-cat+reg}(b) represents the average attribute emotional assessment of valence (1-negative, 5-positive), activation (1-calm, 5-excited), and dominance (1-weak, 5-strong). Audio-alone system has the lowest valence score and highest dominance score among three models, which indicates that audio information is more prone to negative and aggressive, which aligns with the dominant categorical prediction of \textit{angry}. Text-alone system has the lowest activation and dominance score, indicating that text information is more prone to calm and weak. The fused model is relatively positive and excited, in line with the categorical prediction of \textit{happy}. The overall observation against three systems is accordant: depressed samples have higher dominance and activation while lower valence. A possible explanation is that depressed subjects might not reveal their emotions in a direct way -- they might disguise and pretend to be fine. “Smiling Depression” has recently become a rising topic. Besides, in the interview situation, depressed people could be more nervous and their emotion would be amplified while healthy people tend to be more relaxed and calm. This finding is interesting and may be helpful for future psychology study as well.

As shown in Fig.~\ref{fig: mosei-sent}, the sentiment of audio-alone system is slightly biased towards positive while that of text-alone system is biased toward negative. Negative sentiment is discovered after fusing two modalities. In general, depressed samples have higher negative score than healthy samples.
As mentioned in Section~\ref{sec: emo exp}, audio-based emotion features yield worse performance than text-based ones. Combined with the above observation, it is possible that negative emotion is more effective in detecting depression. This also echoes the psychological finding that depression causes increase in habitual use of emotion regulation strategies that serve to down-regulate positive emotion.

To conclude, it is affirmative that depressed mood affects how an individual perceives and expresses emotion. However, according to our results, depression is not always perceived as sad and negative.
Such findings on one hand emphasizes the complexity in depression behavioural signals, but on the other hand lead to a conclusion that emotion can be quite helpful in predicting one's mental state. 
As a matter of fact, not all depressed patients outwardly express emotional symptoms such as sadness or hopelessness \cite{MITCHELL2009609,Schumann_2011}.
We will elaborate individuals' emotion expression analysis in a case-study format. 
\paragraph*{Case Study}
Contradicted emotions from audio and text are sometimes detected from depressed subjects. Two typical depressed samples from DAIC training set are analyzed. Participant 350 was predicted as \textit{happy} by the audio-alone system and \textit{sad} by the text-alone system. From the perspective of speech, the participant has a relaxed tone, an uplifted intonation, and laughter from time to time. 
These traits are strongly correlated with a happy voice and difficult to discriminate from healthy subjects. 
By contrast, when observing the text modality, the overall sentiment is negative filled with destructive words such as ``introvert", ``calm down", ``mistreatment", ``abuse", ``annoyed".
Another sample from participant 426 was predicted as \textit{angry} by the audio-alone system and \textit{sad} by the text-alone system. This participant's speech most qualified as angry: he expressed less and spoke in a downward tone with short finishing. 

As reported, humans can convey inconsistent emotional messages through different modalities \cite{Metallinou_mismatch}. 
We conjecture that emotion expressions carry ample personalized identities, subsequently causing difficulties in clustering into a unified depressive representation.
Emotions are transient and polytropic while depressive mood is more long-lasting and stabilized.
Whilst most people experience depressive symptoms in their life, it is considered an illness, according to the Diagnostic and Statistical Manual of Mental Disorders (DSM) definition \cite{DSM_2013}, when an individual has either a depressed mood or markedly diminished interest or pleasure for longer than a two-week period.
Climate can be inferred from accumulated weather observations however one particular single daily temperature does not necessarily align with the overall profile. 
This echoes our current finding: emotions are helpful in asserting one's mental state nevertheless the relations are not univocal.

\section{Conclusion}
\label{sec: conclusion}
This paper introduces a novel knowledge transfer approach from emotion recognition to depression detection. 
A robust depression detection method is derived from utilizing emotion features extracted from the pretrained emotion model.
In addition, emotion can be seen as an individual modality. We present results fusing emotion with audio and text, elevated at an F1 of 0.87. 

Diversed emotional content of depression data indicates that emotion expressed through audio and text are sometimes inconsistent and negative emotion might be more effective in detecting depression.
The analysis provides clues for understanding the relationship between emotion and depression, in particular how healthy/depressed subjects express their emotions. 
The relationship between emotion and depressed mood is, to some extent, similar to the relationship between weather and climate. 
Emotion expressions are helpful for detecting depression however the relations are complex: depression influences emotions but not simply negative as we expected. 
Depressed patients can exhibit happy emotions just like there are still sunny days in the rainy season.

\bibliographystyle{IEEEbib}
\bibliography{strings,refs}

\end{document}